\documentclass{article}

\usepackage{PRIMEarxiv}

\usepackage[utf8]{inputenc} 
\usepackage[T1]{fontenc}    
\usepackage{hyperref}       
\usepackage{url}            
\usepackage{booktabs}       
\usepackage{amsfonts}       
\usepackage{nicefrac}       
\usepackage{microtype}      
\usepackage{lipsum}
\usepackage{fancyhdr}       
\usepackage{graphicx}       
\graphicspath{{media/}}     
\usepackage{kotex}
\usepackage{amsfonts}       
\usepackage{nicefrac}       
\usepackage{mathtools}

\usepackage{amssymb}
\usepackage{subfig}
\usepackage{tikz}

\usepackage{wrapfig}
\usepackage{algorithm} 
\usepackage{algpseudocode}
\usepackage{multicol}

\title{ Deductive association Networks
}

\author{
  *Seokjun Kim, Jaeeun Jang, *Hyeoncheol Kim \\
  Department of Computer Science and Engineering \\
  Korea University \\
  \texttt{\{*melon7607, wkdwodms0779, *harrykim\}@korea.ac.kr} \\
}

\begin{document}
\maketitle

\begin{abstract}
we introduce deductive association networks, a network that performs deductive reasoning. To have high-dimensional thinking, combining various axioms and putting the results back into another axiom is necessary to produce new relationships and results. For example, it would be given two propositions: "Socrates is a man." and "All men are mortals." and two propositions could be used to infer the new proposition, "Therefore Socrates is mortal.".
To evaluate, we used MNIST Dataset, a handwritten numerical image dataset, to apply it to the group theory and show the results of performing deductive learning.
\end{abstract}

\keywords{Artificial Association Networks \and The second paper \and More experiments are in progress \and More}

\section{Introduction}
Existing neural network models mainly perform from an Inductive Learning perspective over datasets.
The learning method through pattern recognition is a usual inductive learning method. Humans often solve problems from the inductive learning point of view and the deductive learning point of view.
A representative example of deductive reasoning is a syllogism.

\begin{figure}[h]
\centering
\subfloat[Trainset (Proposition)]{
{\includegraphics[width=7.0cm]{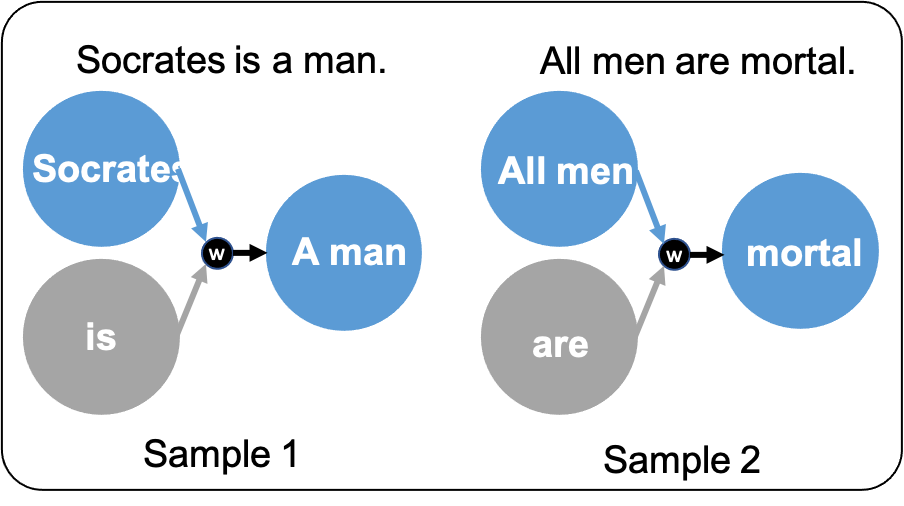}}
}
\subfloat[Testset (Deduction)]{
{\includegraphics[width=5.0cm]{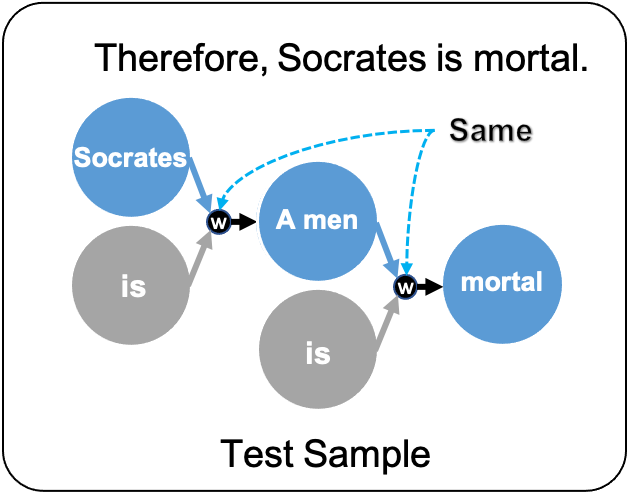}}
}
\caption{we train the proposition or axiom. This network must continuously predict the outcome of all propositions at test time.}
\label{fig:deductive_reasoning}
\end{figure}

To give one example, "Socrates is a man." , "All men are mortal." Two propositions are given, and the new proposition "Therefore Socrates is mortal." can be inferred as Fig.\ref{fig:deductive_reasoning}.
In the human brain, these calculations are typically performed in the prefrontal cortex existing in the frontal lobe, and this reasoning method is widely used in fields such as mathematics and science.
Mathematicians understand axioms and combine them to come up with new principles. Scientists combine real-world phenomena with mathematics and prove them as a result of experiments. In this process, an essential way of thinking is deduction.
In this paper, we propose deductive association networks(DANs), one of the AAN series, designed from this perspective; and to train DANs, we propose the deductive learning methodology.
While the existing neural network models were mainly inductive views that extract and classify features in order to recognize objects, the view of deductive Networks can identify relationships in units of objects, create new propositions, and reach conclusions deductively.
We used the MNIST\cite{lecun1998gradient}, a numerical image dataset, and to perform deductive reasoning, we used group theory\cite{scott2012group} in modern algebra to repeatedly perform a deduction task.
We apply the group theory to show the process of performing deductively successive operations on a finite set with 10 elements and defined operations.
We create episodes of propositions with image datasets, perform Multi-Task to learn propositions, and perform Deductive Learning, which verifies deduction tasks with compound proposition episodes.

\paragraph{Artificial Association Networks}
AAN\cite{kim2021association} is a network introduced in our previous work.
This network consists of a series, and level association networks(LAN) and recursive association networks(RAN) were introduced first, and utilizing these networks.
Association deductive networks is a part of series of AANs that aim to compare with the human brain and present theoretical models and learning methodologies. Because existing networks use fixed layers, they are limited to specific datasets or specific tasks. But in these networks, The propagation structure of existing networks is expressed and learned as a tree structure using a neuro tree data structure without fixed layers.
We want to design a network that can perform tasks that it is difficult for existing neural networks to perform tasks.
This neural network does not learn via fixed layers but proceeds with data-driven learning and encodes as Depth First Convolution (DFC), a type of recursive convolution. It is also possible to decode with the Depth First Deconvolution (DFD) methodology.

The neuro tree(NT) is a data structure that can express relational and hierarchical information.
We can express $\mathbf{NN} = \{x,\tau,\mathbf{A}_c,\mathbf{C}\}$, $\mathbf{NN}_{i} \in \mathbf{NT}$.
The reason for defining the relationship among child nodes is the convenience of implementation.
If we define neuro node(NN) in the above way, we can convert the tree dataset into the NT dataset without significantly modifying the tree structure that has previously been useful.
\paragraph{MNIST Dataset}
MNIST Dataset is an image dataset created by handwriting numbers 0 to 9.
We chose the image dataset to show that our deductive learning theory can be applied to a variety of datasets.
\paragraph{Group Theory}
Group theory\cite{scott2012group} is a field of modern algebra, which is the basis of many fields of mathematics and is to further understand the structure and symmetry of the set by considering the set and operation.
\begin{equation}
G \times G \to G
\label{eqn:grouptheory}
\end{equation}
The definition of the group theory is that if the $\mathbf{(G,\cdot )}$ associative property is valid, the identity element exists, and an inverse element exists for all elements, it is said to be closed for the operation $\cdot$.
The reasons for applying the group theory in this paper are as follows.
"The result of operating with arbitrary elements of the set G exists in the set G again; it is possible to confirm the result with our dataset." And "when a group succeeds in learning, maybe we can infer that all isomorphic groups will succeed in learning."

\section{ Deductive association networks }
\label{sec:network}
In this section, we introduce the architecture of DAN models, one of the AAN series.

\begin{figure}[h]
\centering
{\includegraphics[width=0.20\textwidth]{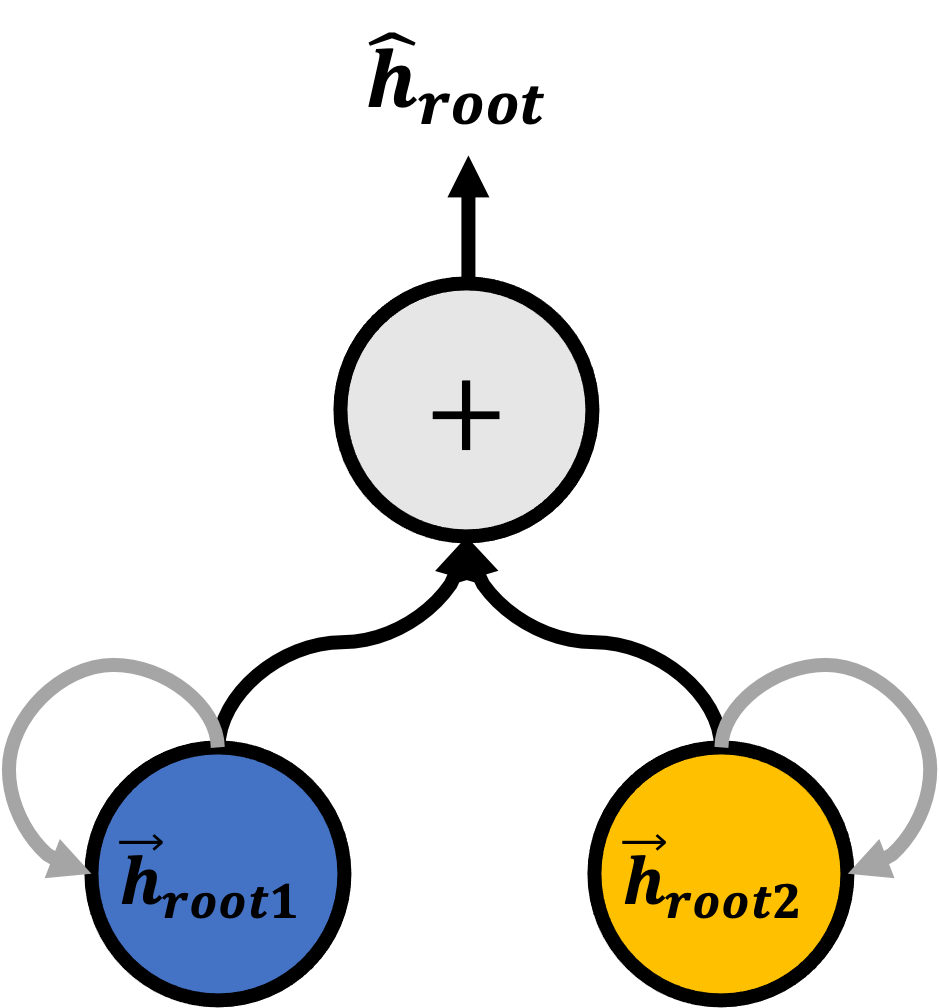}}
\caption{ Neuro tree for deductive reasoning }
\label{fig:neurotree_for_deductive_reasoning}
\vspace{-10px}
\end{figure}

We introduce a neuro tree for performing this deductive reasoning(Fig.\ref{fig:neurotree_for_deductive_reasoning}).
All child nodes are neuro trees for artificial association networks.
And h, the output in the root node, is the input of the deductive model.
And the current node's x contains information about the operation, and its output becomes an input of the deductive model again.

The important point of the network to perform deduction is that "the result of the proposition" should be "used as the input of the next proposition."
DANs are designed with this point in mind. Therefore, there is no hidden vector in this model, and all steps are available as input.
\begin{figure}[h]
\centering
{\includegraphics[width=0.80\textwidth]{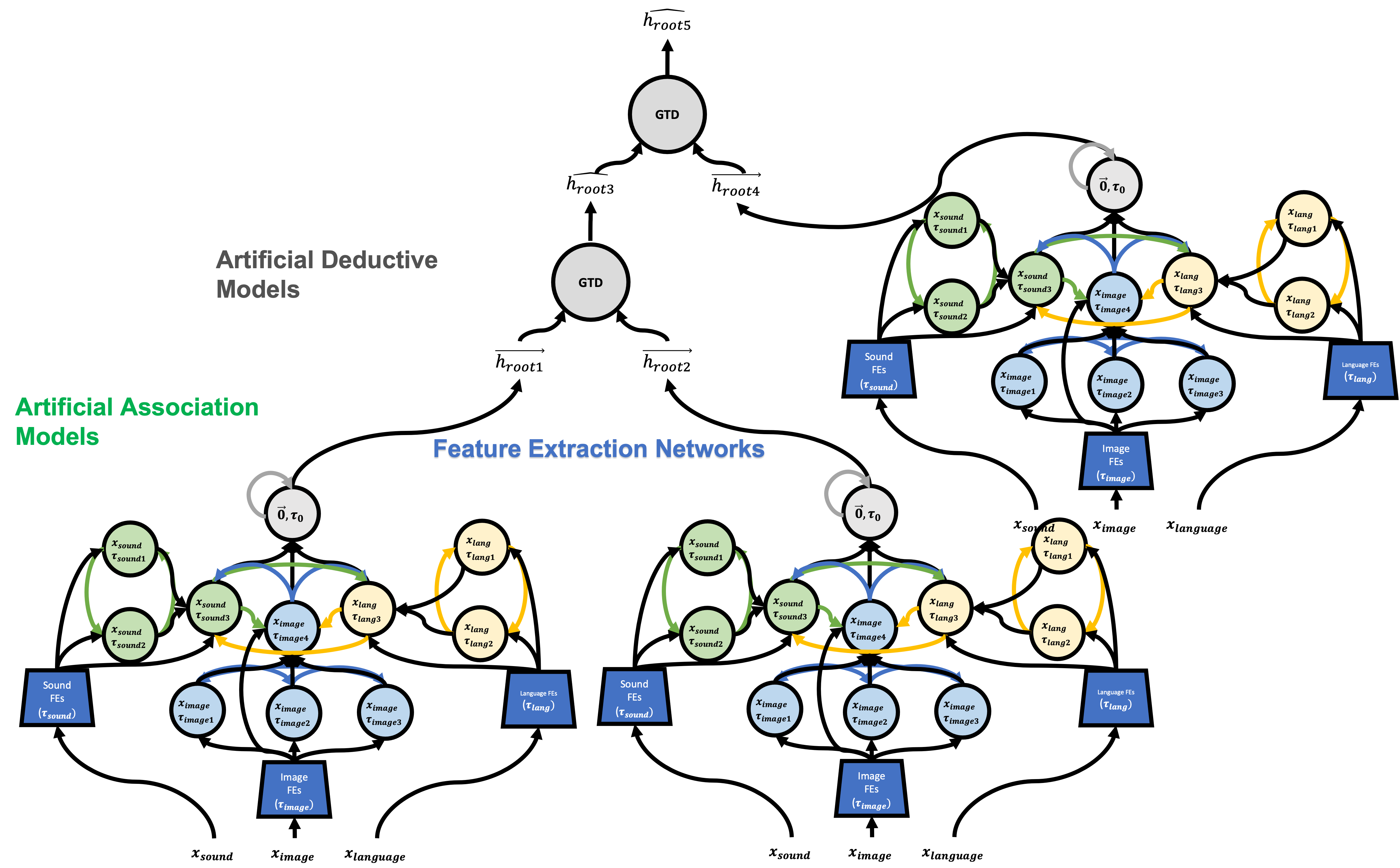}}
\caption{ Ideal Deductive Modeling }
\label{fig:long-term}
\vspace{-10px}
\end{figure}

\label{subsec:gtd}
First, We produce root vectors using LAN or RAN.
We can express the DAN model as $\mathbf{DAN}=\{GRU,\mathbf{W}_{out}, g, \sigma \}$, 
the input and output size of GRU is $F$, and $\mathbf{W}_{out}\in\mathbb{R}^{F \times F}$.
$F$ is the node-feature size and output size. g is readout(max).
\begin{gather}
\overrightarrow{h}_{root_{i}} = RAN(NT)  \label{eq:gtnn}\\[5pt]
\mathbf{E}_{p} = (\overrightarrow{h}_{p1}, \overrightarrow{h}_{p2},... \overrightarrow{h}_{pN}) \label{eqn:gtd_elements}\\[5pt]
\mathbf{x}_{p},\overrightarrow{h}_{hidden}  = GRU(\mathbf{E}_{p}, \overrightarrow{0})\label{eqn:gtdgru}\\[5pt]
\overrightarrow{x'}_{p} = g(\sigma(\tilde{\mathbf{D}}_{p}^{-\frac{1}{2}}\tilde{\mathbf{A}}_{p}\tilde{\mathbf{D}}_{p}^{-\frac{1}{2}}(\mathbf{x}_{p})))\label{eqn:gtdconv}\\[5pt]
\hat{h}_{root} = \sigma([\overrightarrow{x'}_{p}, \overrightarrow{op}_{p}]\mathbf{W})\mathbf{W}_{out}
\label{eq:gtd_out}
\end{gather}

In DANs, the artificial association model encodes the neuro tree and performs deduction using the resulting vector $\overrightarrow{h}_{root}$ (Eqn.\ref{eq:gtnn},\ref{eqn:gtd_elements}).
GRU\cite{chung2014empirical} is used to learn order information(Eqn.\ref{eqn:gtdgru}).
the input information proceeds convolution as Eqn.\ref{eqn:gtdconv}, and $\mathbf{E}_{p}$ means ordered inputs of $\overrightarrow{h}_{root_{i}}$.
The propagation for AAN is proceeding by the DFC traversing from leaf nodes to a root node.
$p$ is the order of DFC, $q$ is the node number of $\mathbf{A}_{c}$, and $N_{p}$ is the number of nodes in $\mathbf{A}_{c}$ of $\mathbf{NN}_{p}$.
$\mathbf{NN}_{p}$ receives $\overrightarrow{h_{pq}}$ from child node $\overrightarrow{h_{pq}} \in \mathbb{R}^{F}$.
we express as $\tilde{\mathbf{A}}=\mathbf{A} + \mathbf{I}$, where $\mathbf{I}$ is the Identity matrix.
If we express the connection value as 1, just the more connected nodes becames the scale value larger than others\cite{kipf2016semi,wu2020comprehensive}.

Therefore, $\tilde{\mathbf{D}}^{-\frac{1}{2}}\tilde{\mathbf{A}}\tilde{\mathbf{D}}^{-\frac{1}{2}}$ is applied using the order matrix $\tilde{\mathbf{D}}$ of $\tilde{\mathbf{A}}$ as a method of normalizing the relationship matrix in Eqn. \ref{eqn:gtdconv}.
Then it goes through the FCNN layer and becomes the final output Eqn.\ref{eq:gtd_out}. $\overrightarrow{op}_{p}$ means operation vector.

And the output of the network can be used as input to the next level convolution.

\subsection{ Deductive Attention Networks }

\label{subsec:gtdan}
We introduce deductive attention networks(DANs) that learn the importance through the attention layer by modifying the expression of the DANs.
It is composed of $\mathbf{DANs}=\{GRU, \mathbf{W}_{out}, g, \sigma, \overrightarrow{a} \}$, and $\overrightarrow{a}$. A parameter for attention mechanism is added ($\mathbb{R}^{2F} \times \mathbb{R}^{2F} \rightarrow \mathbb{R}$).
$\mathcal{N}_{pq}$ is a set of nodes connected to the q-th node in the graph of $\mathbf{NN}_{p}$.
we used LeakyReLU\cite{xu2015empirical} for activation and we have expressed this as:
\begin{equation}
\label{eq:attention_matrix}
\alpha_{pqr} = \frac{\mathbf{exp}(LeakyReLU(\overrightarrow{\mathbf{a}}^{T}[\overrightarrow{x}_{pq},\overrightarrow{x}_{pr}]^{T}))}{\sum_{k\in \mathcal{N}_{pq}}\mathbf{exp}(LeakyReLU(\overrightarrow{\mathbf{a}}^{T}[\overrightarrow{x}_{pq},\overrightarrow{x}_{pk}]^{T}))}
\end{equation}
Through the attention methodology introduced in GATs\cite{velivckovic2017graph} we learn how the r-th node is of importance to the q-th node.
This information is replaced by the connected part of the adjacency matrix.
Therefore, the DANs are as follows:
\begin{equation}
\label{eqn:attention1}
\overrightarrow{x}'_{p} = g(\sigma((\mathbf{A}_{p} \odot \mathbf{\alpha}_{p})(\mathbf{x}_{p})))
\end{equation}
Here, $\odot$ indicates point-wise operation.

To further stabilize the self-attention process, a muti-head attention mechanism is introduced:
\begin{equation}
\label{eqn:attention3}
\mathbf{x}^{k}_{p}, \overrightarrow{h}^{k}_{hidden} = GRU^{k}(\mathbf{E}_{p}, \overrightarrow{0})
\end{equation}

\begin{equation}
\label{eqn:attention3}
\overrightarrow{x}'_{p} = {g(\sigma(\frac{1}{K}\sum^{K}_{k=1}(\mathbf{A}_{p} \odot \mathbf{\alpha}^{k}_{p})(\mathbf{x}^{k}_{p})))}
\end{equation}
where $K$ is the number of multi-heads. The results from multiple heads are averaged and delivered to a parent node. 
Now, this process becomes a cell and delivers the result from the leaf node to the root node by DFC Finally, we get the $\hat{h}_{root}$.

\begin{equation}
\label{eqn:attention1}
\hat{h}_{root} = \sigma([\overrightarrow{x'}_{p}, \overrightarrow{op}_{p}]\mathbf{W})\mathbf{W}_{out}
\end{equation}

\subsection{ Deductive Learning }
Deductive Learning is a methodology that trains propositions and then finds out new compound propositions with the learned network during testing.
This methodology has three requirements.
\begin{itemize}
\item [\romannumeral 1] 
This learning method requires a network that extracts a feature vector as an input object for a proposition.
this network must be jointly training by classification or clustering tasks or use a pre-trained model.
\item [\romannumeral 2] It composes and learns the proposition Episode composed of Clustered Object.
\item [\romannumeral 3] The result of the proposition should be used as input to the next proposition in testing time.
\end{itemize}
This section introduces the training process, and testing process of deductive learning used in this paper.
\begin{figure}[h]
\centering
\begin{tabular}{cc}
\parbox{10.0cm}{~\\[2.8mm]
\rule{0pt}{1ex}\hspace{2.24mm}\includegraphics[width=10.0cm]{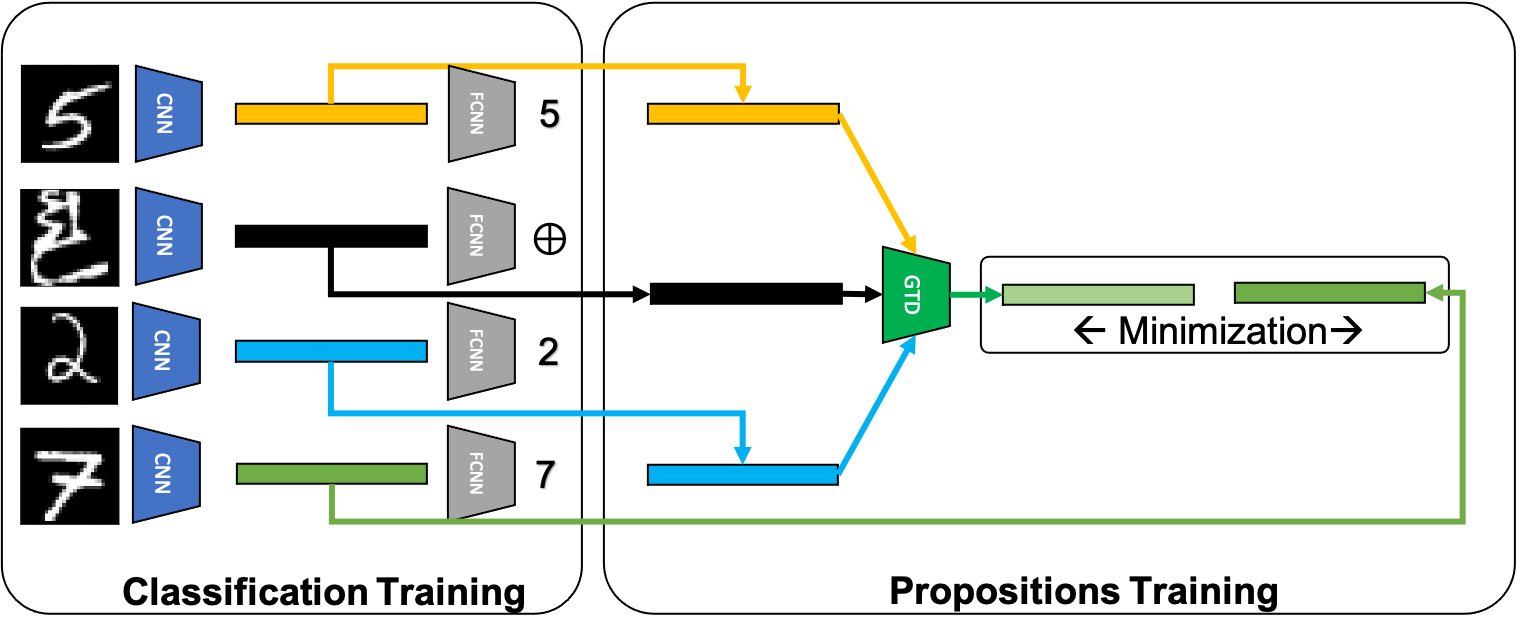}\\[-0.1pt]}
\end{tabular}
\caption{ Deductive Training (Multi-Task) }
\label{fig:deductive}
\end{figure}
\subsubsection{ Training 1 : Classification or Clustering (Recognition task)}
In this step, we perform the task of classifying the object that is the input of the proposition.
The task on the left of Fig.\ref{fig:deductive} belongs to this process.
In our experience, if only the Proposition Episode Training proceeds without this process in learning, the vectors of all objects are learned to be similar vectors.  This process could be replaced by the clustering task.
\begin{equation}
\label{eqn:classification1}
\hat{y}_{q} = Softmax(FC\ layer(\overrightarrow{h}_{root}))
\end{equation}
\begin{equation}
\label{eqn:classification2}
loss_{task1} = \Sigma_{q=0}^{N_{p}}CE(\hat{y}_{q},y_{q})  
\end{equation}
Therefore, we perform Eqn.\ref{eqn:classification1} to make a vector with the same size as the number of classes by FC Layer\cite{minsky2017perceptrons} and then optimize through Cross-Entropy Loss Eqn.\ref{eqn:classification2}.

\subsubsection{ Training 2 : Proposition Episode (Prediction task)}
In this stage, the network is training the propositions.
When feature embedding is performed for each object in training 1, this step is learning to be similar to the target vector of proposition episode by receiving objects as input (Eqn.\ref{eqn:episode1}).
At this time, Episode created by primary propositions using the neuro tree data structure.
This process is on the right side of Fig.\ref{fig:deductive}.
\begin{equation}
\label{eqn:episode1}
loss_{task2} = MSE(\hat{h}_{root},\overrightarrow{h}_{target}) 
\end{equation}
Multi-task learning that performs these tasks jointly is the deductive learning process we used as Eqn.\ref{eqn:episode2}.
\begin{equation}
\label{eqn:episode2}
loss = loss_{task1} + \lambda loss_{task2}
\end{equation}

\subsubsection{ Testing : Compound proposition Episode }
\label{sec:deductive_testing}
A new proposition from which propositions are grouped is called compound proposition.
To validate the trained DAN networks, we use the neuro tree data structure to combine propositions and create deeper test episodes.
The verification method determines whether the class of the output generated by the deductive association network is the same as the target class at the level of all neuro trees.
This network aims to understand simple relationships and principles between objects and to make deductive relationships and principles.
Therefore, we used the output of the proposition existing in each level as the input of the proposition of the next level, continuously determining whether the results of all propositions are the same as the target result, and used it as a verification method for deductive reasoning.
\section{Experimental Results}
We used the MNIST dataset, which is a handwritten numeric image dataset, to show that this structure can be used in various datasets.
Then, the deductive reasoning task was performed to define the operation by applying the group theory and perform deductive reasoning.
\begin{figure}[h]
\centering
\begin{tabular}{cc}
{\parbox{7.0cm}{~\\[2.8mm]
\rule{0pt}{1ex}\hspace{2.24mm}\includegraphics[width=7.0cm]{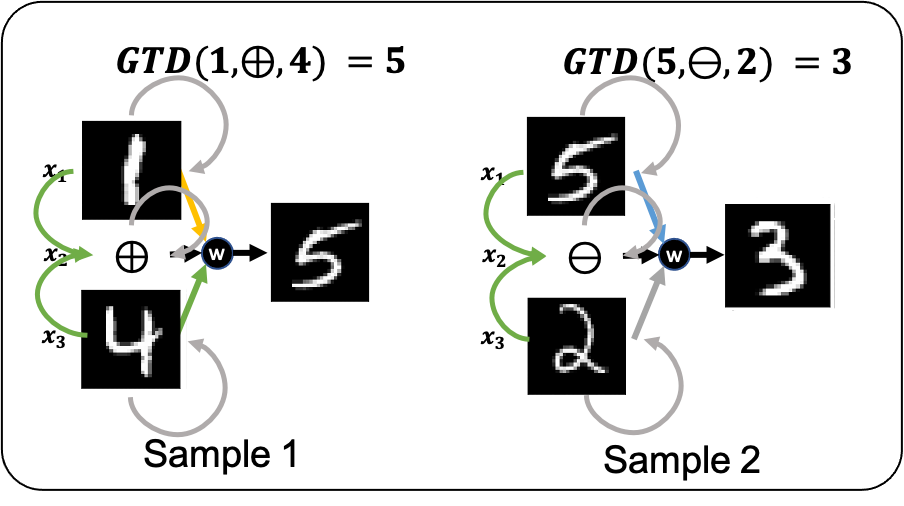}\\[-0.1pt]}} & {\parbox{5.0cm}{~\\[2.8mm]
\rule{0pt}{1ex}\hspace{2.24mm}\includegraphics[width=4.5cm]{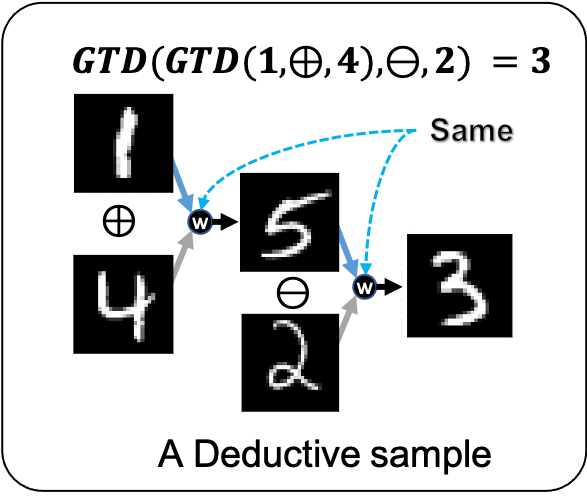}\\[-0.1pt]}}\\
(a) Trainset (Proposition)&(b) Testset (Deduction)
\end{tabular}
\caption{(a) Propositions training; (b) Compound Propositions test }
\label{fig:MNIST_DEDUCTION}
\end{figure}

\subsection{Task 1 : $\oplus$}
\begin{table}[h]
\begin{center}
\begin{tabular}{|l|llllllllll|ccccccccccc|}
\hline
$\oplus$ & 0 & 1 & 2 & 3 & 4 & 5 & 6 & 7 & 8 & 9\\
\hline
0 & 0 & 1 & 2 & 3 & 4 & 5 & 6 & 7 & 8 & 9\\
1 & 1 & 2 & 3 & 4 & 5 & 6 & 7 & 8 & 9 & 0\\
2 & 2 & 3 & 4 & 5 & 6 & 7 & 8 & 9 & 0 & 1\\
3 & 3 & 4 & 5 & 6 & 7 & 8 & 9 & 0 & 1 & 2\\
4 & 4 & 5 & 6 & 7 & 8 & 9 & 0 & 1 & 2 & 3\\
5 & 5 & 6 & 7 & 8 & 9 & 0 & 1 & 2 & 3 & 4\\
6 & 6 & 7 & 8 & 9 & 0 & 1 & 2 & 3 & 4 & 5\\
7 & 7 & 8 & 9 & 0 & 1 & 2 & 3 & 4 & 5 & 6\\
8 & 8 & 9 & 0 & 1 & 2 & 3 & 4 & 5 & 6 & 7\\
9 & 9 & 0 & 1 & 2 & 3 & 4 & 5 & 6 & 7 & 8\\
\hline
\end{tabular}
\end{center}
\caption{ Task 1 diagram of group theory ($\oplus$) }
\end{table}

We selected the group theory task of modern algebra to show the deduction process with the MNIST dataset, and the task performed in this paper is the set G with $\mathbf{G} = \{0,1,2,3,4,5,6,7,8,9\}$ is defined to use a finite group. The operation is described as $\oplus$ and defined as Eqn.\ref{eqn:grouptheory1}.

\begin{equation}
G\oplus G\to G
\label{eqn:grouptheory1}
\end{equation}
\begin{equation}
x_{1} \oplus x_{2} = (x_{1} + x_{2})\ mod\ 10
\label{eqn:oplus}
\end{equation}
This operation adds two elements and performs a modulo operation as Eqn.\ref{eqn:oplus}.
And the identity element is 0, the inverse element for $\oplus$ of elements of G is always present, and the associative property is valid. Therefore, this operation satisfies $\mathbf{G} \oplus \mathbf{G} \to \mathbf{G}$.
Therefore, the image corresponding to the elements of $\mathbf{G} = \{0,1,2,3,4,5,6,7,8,9\}$ is recognized as one element and there are 10 elements. Define finite group, perform deduction task, and show experimental results.

\subsection{Task 2 : $\ominus$}
\begin{table}[h]
\begin{center}
\begin{tabular}{|l|llllllllll|ccccccccccc|}
\hline
$\ominus$ & 0 & 1 & 2 & 3 & 4 & 5 & 6 & 7 & 8 & 9\\
\hline
0 & 0 & 1 & 2 & 3 & 4 & 5 & 6 & 7 & 8 & 9\\
1 & 9 & 0 & 1 & 2 & 3 & 4 & 5 & 6 & 7 & 8\\
2 & 8 & 9 & 0 & 1 & 2 & 3 & 4 & 5 & 6 & 7\\
3 & 7 & 8 & 9 & 0 & 1 & 2 & 3 & 4 & 5 & 6\\
4 & 6 & 7 & 8 & 9 & 0 & 1 & 2 & 3 & 4 & 5\\
5 & 5 & 6 & 7 & 8 & 9 & 0 & 1 & 2 & 3 & 4\\
6 & 4 & 5 & 6 & 7 & 8 & 9 & 0 & 1 & 2 & 3\\
7 & 3 & 4 & 5 & 6 & 7 & 8 & 9 & 0 & 1 & 2\\
8 & 2 & 3 & 4 & 5 & 6 & 7 & 8 & 9 & 0 & 1\\
9 & 1 & 2 & 3 & 4 & 5 & 6 & 7 & 8 & 9 & 0\\
\hline
\end{tabular}
\end{center}
\caption{ Task 2 diagram of $\ominus$ }
\label{table:task2}
\end{table}
\begin{equation}
G\ominus G\to G
\label{eqn:grouptheory2}
\end{equation}
\begin{equation}
x_{1} \ominus x_{2} = (x_{1} - x_{2})\ mod\ 10
\label{eqn:ominus}
\end{equation}
This operation subtracts two elements and performs a modulo operation.
However, the operation $\ominus$ does not satisfy the group (a $\ominus$ e = e $\ominus$ a = a).
But, like table.\ref{table:task2}, $\mathbf{G} \ominus \mathbf{G} \to \mathbf{G}$ is proven, therefore we tested it with Task 2.

\subsection{Task 3 : $\oplus$, $\ominus$}
Both operations in Task 1 and Task 2 use the same set of elements, and $\mathbf{G} \ominus \mathbf{G} \to \mathbf{G}$ is proven. Therefore, it is possible to learn these two operations together. Testing them together is Task 3, and Fig.\ref{fig:MNIST_DEDUCTION}.
Here, operator input is added, and an operator is selected randomly. And two images of KMNIST were added and replaced to mean an operator.

\subsection{Evaluation}
We performed the evaluation of Task 1, Task 2, and Task 3 described above, and the result is Table.\ref{fig:experimental_result}.

\paragraph{Deductive Training}
In our experience, The important thing is that learning is not possible when learning a proposition if each object is not classified. And the percentage of correct answers to proposition increases when object classification goes well. 
This is the reason for applying the multi-task learning methodology.

\paragraph{Deductive Testing}
As mentioned in \ref{sec:deductive_testing}, We applied it to the test dataset. The evaluation method is whether the output of the proposition at all levels matches the target class.
We performed the task of solving the proposition deductively.
The number of deductions becomes depth, and the deeper the neuro tree, the more deductions. Therefore, the deeper the depth, the more complex the task, and the result are as shown in the Table.\ref{fig:experimental_result}
\begin{table}[h]
\begin{center}
\begin{tabular}{|l|l|lllll|c|c|ccccc|}
\hline
episode & proposition & syllogism & sorites & sorites & sorites & sorites \\
\hline
$\times$ & depth 0 & depth 1 & depth 2 & depth 3 & depth 4 & depth 5\\
\hline
Task 1 $\oplus$ & 97.51\% & 95.73\% & 94.57\% & 92.41\% & 91.03\% & 89.56\% \\
Task 2 $\ominus$ & 97.15\% & 95.76\% & 94.19\% & 93.22\% & 91.38\% & 90.28\% \\
Task 3 $\oplus,\ominus$ & 97.71\% & 96.17\% & 95.63\% & 94.07\% & 93.08\% & 91.46\% \\
\hline
\end{tabular}
\end{center}
\caption{ Deductive Testing Results of MNIST Dataset }
\label{fig:experimental_result}
\end{table}

\section{Conclusion}
In this paper, the deductive learning model, one of the artificial association networks model series, was introduced and applied to the image dataset.
This paper is part of a series; in the next paper, we introduce memory association networks(MANs).

\bibliographystyle{splncs04}  
\bibliography{main}

\end{document}